# Understanding with Toy Surrogate Models
# in Machine Learning

Andrés Páez
Universidad de los Andes

### ABSTRACT

In the natural and social sciences, it is common to use toy models—extremely simple and highly idealized representations—to understand complex phenomena. Some of the simple surrogate models used to understand opaque machine learning (ML) models, such as rule lists and sparse decision trees, bear some resemblance to scientific toy models. They allow non-experts to understand how an opaque ML model works globally via a much simpler model that highlights the most relevant features of the input space and their effect on the output. The obvious difference is that the common target of a toy and a full-scale model in the sciences is some phenomenon in the world, while the target of a surrogate model is another model. This essential difference makes toy surrogate models (TSMs) a new object of study for theories of understanding, one that is not easily accommodated under current analyses. This paper provides an account of what it means to understand an opaque ML model globally with the aid of such simple models.

**Keywords**: Toy models, Surrogate models, Machine learning, Understanding, Idealization

## 1. Introduction

In the natural and social sciences, it is common to use extremely simple and highly idealized models to understand complex phenomena. Unlike regular models, these very simple models—often referred to as *toy models*—are not required to be linked to the real world through structural similarity or resemblance relations. They are not meant to be approximations of the target world system, and in some cases, they are not even required to be representational. In semantic terms, they do not accurately map onto their targets. Despite these limitations, they are still useful in understanding theoretical concepts and possible configurations of the target system. Paradigmatic examples of toy models include Boyle's law and the Ising model in physics, the Lotka–Volterra model in population ecology, and the Schelling model in the social sciences (Weisberg, 2013).

In recent years, philosophers of science have become interested in toy models (Grüne-Yanoff, 2009; Luczak, 2017; Reutlinger et al., 2018; Frigg & Nguyen, 2017; Nguyen, 2020). The main purpose of this literature is to explore the nature of these models and examine how they perform their epistemic function. Despite lacking the regular descriptive and predictive features of full-scale scientific models, they often offer an elementary understanding of a phenomenon. Their definitions of "toy model" differ as well as their assessment of the importance of representation in modelling generally, but they all agree that toy models play an important epistemic role in scientific research, exploration, and pedagogy.

*Prima facie*, some of the proxy, interpretative, approximate, or surrogate models[1] used in explainable AI (XAI) to make sense of black box machine learning (ML) systems play an analogous role to toy models in the sciences.[2] In both cases, the models fulfill what Frigg and Nguyen (2020, p. 3), following Swoyer (1991), call the *surrogative reasoning condition* for representation: models represent in a way that allows scientists or users to make inferences about the models' target systems; they can generate claims about target systems by investigating models that represent them. Although many surrogate models used by developers in ML are black boxes,[3] the simplest of them—e.g., rule lists and sparse decision trees—allow non-experts to understand how an opaque ML model works globally via a much simpler model that highlights the most relevant features of the input space and their effect on the output. Toy surrogate models (TSMs), as I will call them, only work when the system's features can be interpreted semantically, that is, when they represent recognizable elements of the user's environment. It is well-known that many ML systems use non-interpretable features that would impede the extraction of a TSM. The examples used in this paper therefore assume that the features are human-interpretable. The ultimate goal of TSMs is to provide the end users of an AI system with valuable understanding that will result in informed decisions and/or actionable changes.

---

[1] I will refer to these models as "surrogate models," but some papers use the other terms to refer to models that perform the same epistemic function.

[2] In this paper, I will assume that the reader is familiar with the problem of opacity in machine learning and with the literature on interpretability and XAI. For an introduction to the topic and some of the controversies involved, see Beisbart and Räz (2022), Humphreys (forthcoming), Krishnan (2020), and Lipton (2018).

[3] For example, Xu et al. (2018) build a surrogate model by compressing an existing DNN model to a shallow DNN model, but the latter is still a black box.

TSMs can be a valuable instrument to comply, for example, with Article 13 of the GDPR (Regulation EU 2016/679) which requires the data controller to provide the data subject with "meaningful information about the logic involved" whenever automated decision-making tools are used.

Despite having similar epistemic roles, the relation between TSMs and opaque ML models is different than the relation between their counterparts in the sciences. Toy models and complex models in the sciences share a common target: some social or physical phenomenon that can be understood either in highly idealized and simple terms through the toy model, or in a more complex and detailed fashion—often involving causation and lawlike generalizations—via the main model. In contrast, the most common use of discriminative ML models is to perform a prediction or classification task that is not necessarily causally grounded in the world or reflective of lawlike relations between inputs and outputs. In other words, most ML models do not have the same representational and epistemic function as the models used in the natural and social sciences. They do not aim at uncovering complex real-world causal or lawlike structures that are responsible for the properties of a phenomenon, but rather to detect useful correlations that optimize the predictive or classificatory task at hand.[4] Toy surrogate models in ML, in turn, focus on the statistical correlations in the main model, which they aim to approximate and present in simpler and understandable terms. They are models of models, i.e., metamodels (Alaa & van der Schaar, 2019).

As I will be discussing models of various types, it will be helpful to explain how models are conceptualized both in science and in ML. In the social and natural sciences, many models are sets of mathematical equations designed to represent specific aspects of the world. However, not all scientific models are mathematical. Some are mechanical artifacts or graphical representations; others are fictional objects, abstract objects, set-theoretic structures, or a combination of all of the above. In a sense, anything can be a scientific model, a fact that has led many to voice skepticism about the prospect of offering a definition of the kind of object they are (Frigg & Hartman, 2020). Toy models in the sciences are equally diverse; the only difference with full-scale scientific models is that they are much simpler, mathematically and structurally. ML models, on the other hand, are always mathematical in nature. Although deep neural networks, for example, are often represented using diagrams, at the computational level they are just extremely complex mathematical functions. Most TSMs are also mathematical structures, but some may combine other elements. Decision trees and rule lists, for example, combine mathematical and logical operations. The relation between TSMs and their targets occurs entirely at the computational level. They do not include any information about the algorithmic or the implementation levels of analysis, to use Marr's (1982) well-known taxonomy. Their purely computational nature makes them akin to the toy mathematical models used in the sciences, In both scientific and ML contexts, toy models play the same epistemic role of helping us understand how full-scale models perform their representational or discriminatory task, as I argue in section 3.

---

[4] There are, of course, significant efforts to use ML models in scientific research to uncover causal relations in the world from large datasets (Pearl, 2019) and to implement causal models (Buijsman, 2023; Pietsch, 2016; Sullivan, 2023). In this paper I am concerned mostly with discriminatory ML systems designed for more practical predictive and classificatory tasks.

Some readers might wonder about the place of computer simulations (CS) within this discussion.[5] CS are quite distinct from ML models; they follow different principles, have different properties, and are applied to different problems (Alvarado, 2023). Nonetheless, there are interesting intersections between CS and ML models. On the one hand, ML models are used sometimes to emulate the behavior of computationally expensive CS (Angione et al., 2022). In these cases, it seems more natural to call the surrogate ML models meta-simulations because the models are trained on the simulations' reaction to a restricted number of carefully selected data points. On the other hand, CS have been used to support surrogative reasoning in AI-based engineering (Datteri & Schiaffonati, 2023). In both cases, the surrogate and the target are opaque structures. The kind of surrogate models I discuss in this paper are not and cannot be black boxes if they are to fulfill the epistemic purpose of helping lay users understand opaque structures.

The literature on CS provides us with a clear definition of the kind of opacity we will be dealing with. In many CS, the relationship between the initial and final steps is epistemically opaque because it is impossible "to decompose the process between model inputs and outputs into modular steps, each of which is methodologically acceptable both individually and in combination with the others" (Humphreys, 2004, pp. 147-148). The epistemic opacity of CS is a property they share with many ML models. More precisely, CS are *essentially* epistemically opaque: "A process is essentially epistemically opaque to X if and only if it is impossible, given the nature of X, for X to know all of the epistemically relevant elements of the process" (Humphreys, 2009, p. 618). Although in this paper I discuss the epistemic role of TSMs in providing some degree of understanding of opaque ML target models, the goal is obviously not to achieve the modular decomposition of the target model, as stated in Humphreys' definition. Understanding will be obtained at the computational level, not the algorithmic level.

---

[5] I am grateful to an anonymous reviewer for inviting me to clarify the relation between computer simulations and ML models. I can only touch on the subject very briefly here.

The main purpose of this paper is thus to explore the analogies and disanalogies between the nature and epistemic role of toy models in the sciences and in ML. More specifically, if scientific toy models provide some degree of understanding of complex natural and social phenomena represented by a full-scale scientific model, what kind of understanding do TSMs provide in the context of ML? Do they provide a similar kind of understanding, or do they play a unique epistemic role given the unique nature of their target? Part of the answer lies in the kind of relation that holds between surrogate and target models. Most theories of scientific models understand their representational character in terms of isomorphism (van Fraassen, 1980), analogy (Hesse, 1966) or similarity (Giere, 1988; Weisberg, 2013) to their targets. As I explain in Section 4, none of these kinds of relationships are achievable in ML. The alternative is to adopt a version of a pragmatic theory of representation (Giere, 2004; Suárez, 2004). These theories expand the dyadic relation between the model and the world by including the users of the model in the picture. Representation in this approach is seen as an epistemic accomplishment of the users of the model, and it is dependent on their epistemic goals. We can think of TSMs as epistemic tools (Knuuttila, 2011; Currie, 2017) that help agents understand the target model. The degree of understanding they achieve can be fleshed out in terms of their ability to use and manipulate the target system and make inferences about it. Understanding in ML will turn out to be a sort of pragmatic achievement.

The paper is organized as follows. The first task will be to understand what a global surrogate model is and how it differs from local *post-hoc* interpretation methods (section 2). In section 3, I briefly explore different approaches to toy models in the sciences and whether these models share the main features of TSMs in ML. In section 4, I explore the relation between surrogate and target models. I conclude that the relation has an ineliminable pragmatic component that depends on the purpose and stakeholders of the target model. The final task of the paper will be to inquire how TSMs provide understanding of the original ML model (section 5). It is important to keep in mind that my interest in this paper is not on how ML can be used in scientific research to help us understand some aspect of the world (Sullivan, 2022, 2023; Zednik & Boelsen, 2022), but rather on how toy surrogate models can provide understanding of an opaque ML system for its lay stakeholders.

## 2. Local and Global XAI Methods

XAI methods can be readily classified in terms of their scope into local and global. The former attempt to explain the singular predictions of an ML model, while the latter provide a general idea of the functioning of the system. Another way to put it is that the former methods explain decisions while the latter explain the capabilities of the model. Local interpretations include counterfactual probes (Wachter et al., 2018; Mothilal et al., 2020), and different types of perturbation-based methods such as LIME (Ribeiro et al., 2016;), Grad-CAM (Selvaraju et al., 2017; Ancona et al., 2019), SHAP (Lundberg & Lee, 2017), TCAV (Kim et al, 2018), among many others (see Ivanovs et al., 2021 for a survey).

Global explanations, on the other hand, are based on proxy, interpretative or surrogate models. They are lighter versions of the original model, trained to mimic the behavior of the original model. More precisely, given a model $f$, the goal is to generate a simpler model M such that $M(x) \approx f(x)$. Surrogate models deliver faster predictions while using less processing power and memory. Some surrogate models are generated through knowledge distillation techniques (Kim et al., 2022; Tan et al, 2018). The original model $f$ is called the teacher model, and the distilled one the student model. Some of the resulting models are understandable for non-experts and are thus toy surrogate models. The most widely used classes of TSMs are linear or gradient-based approximations, decision rules, and decision trees (Frosst & Hinton, 2017; Wu et al., 2018). To create a manageable linear model, expert knowledge is required to select the features that will be included. Only features that exceed a certain threshold of correlation between the feature and the target should be used, but there is always the risk that some features might not show an individual correlation, and/or that their contribution only becomes clear in combination with other features. The advantage of linear models is that they are widely used in the natural and social sciences, including medicine, which makes them a familiar and accepted tool for many of their intended users. Decision trees are used in cases where the relation between features and targets is nonlinear or where features interact with each other. They can also be expressed as decision rules. However, their step-by-step nature is not very efficient. They are also very sensitive to small changes in the training dataset or to a change in feature choice: a change in a split high up the tree will affect the entire tree. Finally, decision trees can also become unmanageable because they grow exponentially, which raises the question: at what point do they cease to be toy models? The answer has to be reached empirically through user studies, as I argue below. Examples of TSMs include the weighted checklists proposed by Jung et al. (2017), the decision rules presented in Letham et al. (2015), and the decision trees for diabetes risk prediction introduced by Bastani (2019), which can be readily used by the judicial and medical professionals for whom they are intended.[6]

We will examine some of these techniques in more detail in the next section when we compare them to scientific toy models. In the rest of this section, I want to defend the claim that the limitations of local *post hoc* interpretability methods make them a very imperfect tool for understanding ML models, and that global surrogate models offer a better epistemic approach to the problem of opacity in ML. In section 5, I will also show that, from an epistemological point of view, global methods are needed to make sense of the results of local methods.

During a long time, local *post-hoc* interpretability methods were seen—at least within the community of AI developers—as the most promising approach to open the black box of AI. Recently, however, they have become the subject of much criticism due to their intrinsic limitations and weaknesses, and because of the inscrutability of the resulting "explanations" for non-expert end-users and stakeholders (Ehsan & Riedl, 2020; Rudin, 2019). Perhaps the most damning problem for local methods is that they perform poorly on diverse robustness metrics. Ideally, very similar inputs should not generate substantially different explanations. But simple transformations of the input, or repeating the sampling process, can generate different explanations. Kindermans et al. (2019) show that adding a constant shift to the input data, which is a simple and common pre-processing step with no effect on the performance of the model, causes numerous interpretation methods to make incorrect attributions. Slack et al. (2020) show the vulnerability of LIME and SHAP to adversarial attacks. And Ghorbani et al. (2019) demonstrate how to generate adversarial perturbations that produce perceptively indistinguishable inputs that are assigned the same predicted label yet result in very different interpretations using feature importance methods.

---

[6] A different approach, which will not be discussed in this paper, is to use example-based methods (Kenny et al., 2021). The idea is to select subsets of the dataset to explain the behavior of ML models or to make explicit the underlying data distribution. This approach only works when the data is structured and can be represented in a human understandable way. They include similar examples or factuals (Schoenborn et al., 2021), influential instances (Koh et al., 2017, 2019) and prototypes (Kim et al., 2016), among others. These methods yield results that are more easily understood by non-experts, but they are prone to misinterpretation, as Sara Mann (2024) has shown.

Another limitation of local methods such as heat and saliency maps is that they lack precision. Rajpurkar et al. (2017), for example, use a heatmap method that highlights the areas of an x-ray deemed most important for the diagnosis of pneumonia by a 121-layer convolutional neural network. Ghassemi et al. (2021) argue that even the hottest parts of the map contain both useful and useless information (from the perspective of a human expert), and simply localizing the region does not reveal exactly what the model considered important in that area. Furthermore, the information provided in the hot area has to be interpreted, thereby opening the door to the clinician's previous beliefs and the risk of confirmation bias. The explanation also lacks any sort of justification of why that particular area was more relevant than others because there is no causal knowledge supporting the explanation. Finally, automation bias (Lyell & Coiera, 2017) can lead to an overestimation of the ML system's performance.

Counterfactual methods are not immune to the robustness problem. Like perturbation-based methods, counterfactual methods can be manipulated and may converge towards drastically different explanations under small perturbations (Slack at al., 2021). Counterfactual probes also critically depend on closeness metrics but there is no principled way to decide which metric to use in any given case. And like saliency-based methods, the lack of causal grounding can deliver sub-optimal or even erroneous explanations to decision-makers (Chou et al., 2022).

This is just a small sample of the problems faced by local explanations. Their fragility and imprecision are grave enough to recommend at least combining them with global methods to achieve a better understanding of ML systems. To be sure, global explanation methods are not a panacea. Many global methods do not present the difference in prediction performance and feature contribution according to the degree of distillation of the student model, which is an important measure of model simplification (Kim et al., 2022). Also, many surrogate models that are easy to understand incur in overfitting and deteriorated accuracy compared to the teacher model. Of course, one can argue that the main goal of a TSM is not to achieve a similar level of performance to the original model, but rather to help users obtain a basic understanding of how it works at a computational level. Simple explanations might help users understand the capabilities and limitations of the model and adjust their confidence levels accordingly. Simple decision trees, rule lists, example-based methods, and even dialogical explanations will perform much better in this kind of task. The nature of some of these simpler surrogate models is the topic of the next section.

## 3. Toy Models in the Sciences and in Machine Learning

In this section we will take a closer look at the nature and epistemic role of toy models in the social and natural sciences so we can compare them to toy surrogate models in ML. There are at least three different descriptions of toy models in the sciences. We will examine them in turn. On one end of the spectrum, we find *targetless* toy models. A first approach to these models, championed by Grüne-Yanoff (2009) in the field of economics, conceives of toy models as devices that present a possible configuration of the world that contradicts or questions an impossibility result. Showing that a toy model is possible can affect one's confidence about what is impossible or necessary in the real world. It is not important that toy models satisfy "world-linking" properties such as being similar or isomorphic to any real target, or that they reflect regularities in the world, because their purpose is purely theoretical: to show a possible world with some variables of interest. More generally, targetless toy models can be used in hypothetical thinking, for example, by trying to establish the necessary conditions of existence of non-existing phenomena. For example, a complete understanding of a biological process "requires learning about the properties of both actual and nonactual target systems" (Weisberg, 2013, p. 121).

These targetless models need not concern us here because surrogate models in ML have a concrete target. To be sure, in XAI sometimes we are interested in understanding how a model might have produced a different outcome if the input had been different. This might seem like an attempt to understand a nonactual target, but the parallel with hypothetical thinking is illusory because the model itself remains unaltered and continues to be the static target of the surrogate model.

A second type of scientific toy model is introduced, not with the purpose of understanding an individual phenomenon but rather classes of phenomena, or "generalized targets," as Weisberg (2013, p. 114) calls them. He mentions, for example, parasitism, sexual reproduction, evolution, and chemical bonding as phenomena that can be modelled without targeting any specific instance of them. Although a model might be used to explain how a type of ML algorithm such as a DNN works in general terms, my main concern in this paper is about the use of TSMs to understand ML models that have been trained to perform specific tasks.

The most common characterization of scientific toy models—and the one we will focus on in the rest of this paper—imposes the condition that toy models must refer to or represent a concrete target system (Reutlinger et al., 2018; Nguyen, 2020). The main point of contention, of course, is what is meant by *representation* in this context. I will return to this point in the next section. All authors also assert that toy models must be very simple, but simplicity comes in degrees, and everyone agrees that there is a continuum between toy models and regular models. The same is true about the degree of idealization involved. Clear cases of toy models simply occupy one end of the spectrum. Reutlinger et al. (2018) defend the additional claim that the main epistemic role of toy models is to provide individual scientists with *understanding* of the target phenomenon. Being literally false representations of their target phenomena, their epistemic goal is neither prediction nor explanation, at least if explanation is understood in factive terms (Páez, 2009). Some philosophers argue that genuine understanding can only be achieved if models provide a mostly truthful representation of their target phenomena. This is the position defended by *quasi-factivists* (Greco, 2014; Khalifa, 2012, 2017; Kvanvig, 2003, 2009; Mizrahi, 2012). In Kvanvig's view, for example, idealized models only offer understanding "in an honorific sense" (Kvanvig, 2009, p. 341); they do not provide bona fide scientific understanding. In contrast, *non-factivists* like Catherine Elgin reject this requirement and settle for idealized models as "felicitous falsehoods" (Elgin, 2009, p. 327).[7] In her words, we must only require idealized models to be "true enough" of the phenomena, but whether they are true enough is clearly a contextual question (Elgin, 2008, p. 85). I will postpone discussion of the relation between factivity and understanding in ML until the next section of the paper, but it should be obvious that my sympathies lie in the non-factivist camp.

It should be clear from the discussion in this and the previous section that not all surrogate models in ML qualify as toy models in the sense used in the social and natural sciences. Given the well-known tradeoff between precision and interpretability, any surrogate model that pretends to preserve a high degree of precision in comparison with the original model will most probably fail the simplicity test. Admittedly, we are navigating in the dark here because there is very limited evidence about the comparative understandability of methods such as single-hit decision tables, binary decision trees, propositional rules, oblique rules, sparse linear regressions, and simple naïve Bayes classifiers. There is an urgent need to test these methods in terms of accuracy, response time, and answer confidence for sets of problem-solving tasks on a diverse population with different levels of expertise (Doshi-Velez & Kim, 2017).[8] To complicate things, there seems to be a lack of interest in the AI community to test whether the XAI methods they employ are actually understood by non-experts. A recent survey shows that only 36 out of 117 (31%) research papers evaluating counterfactual explanations included user studies (Keane et al., 2021). One cannot help but agree with the authors when they say that "the XAI community is busily developing technical solutions that may have no practical benefits to people in real-life" (Keane et al., 2021, p. 1).

This leaves us with only two methodological choices to investigate TSMs: either to stipulate some desiderata that surrogate ML models must fulfill to be considered toy models, or to find some paradigmatic examples of toy models in ML. The first approach would be highly arbitrary given the dearth of evidence in support of specific features that increase the understandability of a model. This leaves us with the option of analyzing paradigmatic examples of very simple surrogate methods. The use of paradigmatic examples is common in the philosophical literature on scientific toy models, so it should not be seen as an unusual choice in this context. The examples I will use are the decision rules presented in Letham et al. (2015), the checklists proposed by Jung et al. (2017), and the decision trees for diabetes risk prediction introduced by Bastani et al. (2017). I will not consider example-based methods because there is a separate philosophical literature on the understanding provided by examples (Elgin 2016, 2017; Khalifa, 2017; Mann, 2024) and analogies (Bartha, 2010; Coll et al., 2005; Hesse, 1966). Examining this literature would take us too far afield.

---

[7] See also Doyle et al. (2019), De Regt (2015), Elgin (2004, 2007, 2008, 2017), and Potochnik (2017, 2020).
[8] The very short list of experiments in this field include: Allahyari and Lavesson (2011), Freitas (2014), Fürnkranz et al. (2018), Lage et al. (2019), Kandul et al. (2023), Kliegr et al. (2018), Piltaver et al. (2016), and van der Waa (2021).

Letham et al. (2015) propose interpretable prediction models that take the form of sparse decision lists. Each list consists of a series of association rules in the form of if-then statements. The rules are pre-mined from the input space using the FP-Growth algorithm (Borgelt, 2005). The resulting models have the same level of complexity as medical scoring systems, thus making them easy to use for clinical practitioners. In terms of performance in a stroke risk classification task, the decision lists were comparable to a support vector machine (SVM), and not substantially worse than an $L^1$ logistic regression and a random forest trained on the same data.

Jung et al. (2017) present a method for constructing simple rules to make complex decisions. The rules take the form of a weighted checklist that can be applied mentally and equals the performance of opaque ML algorithms such as random forests. The authors describe their approach thus:

> Our *select-regress-and-round* strategy results in rules that are fast, frugal, and clear: fast in that decisions can be made quickly in one's mind, without the aid of a computing device; frugal in that they require only limited information to reach a decision; and clear in that they expose the grounds on which classifications are made (Jung et al., 2017, p. 1).

In a nutshell, the technique *selects* a (usually small) subset of features which are then used to train an $L^1$ regularized logistic *regression* model. The coefficients are then rescaled within a restricted range and *rounded* to the nearest integer. Their technique is demonstrated in the case of judicial decisions to release or detain defendants while they await trial. Using a dataset of 165,000 cases, decision rules using only two features—age and previous failure to appear in court—were on a par with decisions derived from random forests trained on 64 available features. Although these two features were selected because they are independently known to be highly predictive in this context, the authors show how such features can be selected in a principled fashion without domain expertise. The method was generalized to 22 other decision-making domains, where it showed to be competitive with opaque ML algorithms.

Bastani et al. (2017) devised an algorithm for extracting simple decision tree explanations from complex models. The method was tested on two tasks, only one of which need concern us here. The test model was a black box random forest trained to classify patients as high or low risk for type II diabetes based on their prescribed medications, demographics, and diagnosis codes, as specified in the 9[th] edition of the *International Classification of Diseases* (ICD-9). A binary decision tree was extracted from the model and tested for fidelity and interpretability. Although the tree performs better than other simple methods trained on the same test set—including the rule lists presented in Letham et al. (2015)—its performance was worse than the original model, as was to be expected. The tree's interpretability was established through a questionnaire that required subjects to perform tasks such as computing counterfactuals and identifying risky subpopulations. They could use either the decision tree or a rule list extracted using the method introduced in Yang et al. (2017). Subjects in the user study were able to answer correctly more questions about risky populations using the decision tree than using the rule list. As Lakkaraju et al. (2016) have shown, decision trees in general are easier to understand than rule lists.

After examining these examples, their similarity to the toy models used in scientific contexts is not self-evident. In particular, it is not clear whether the TSM *represents* the target model in the same way that a toy scientific model represents the same aspect of the world as the full model. To address this question, in the next section I will examine several ways in which the relation between surrogate and target models can be understood.

## 4. The Relation between Surrogate and Target Models

In this section, I will examine the possible ways in which the relation between surrogate and target models can be conceptualized. In general terms, there are two different paths that can be taken. The first one is to examine the *informational* and *structural* relation between the models. The second one is to focus on the *pragmatic* aspects of that connection. The main difference between these perspectives is that the former is a two-place relation between model and target, while the latter is a three-place relation between model, target, and user of the model. We will examine each one in turn.

From an informational and structural perspective, an initial possibility is to think of surrogate models as *idealizations* of their targets. After all, toy models in the sciences are highly idealized representations. The explanation of why gases behave according to Boyle's law is based in the ideal gas model in which molecules are infinitely small and never collide. Models in fluid dynamics also assume infinitely small particles, and models in population genetics assume infinitely large populations without genetic drift (Strevens, 2017).

It is customary to distinguish between two main types of idealizations, Galilean and Aristotelian (Frigg & Hartmann, 2020; Weisberg, 2007) Galilean idealizations are distortions and omissions introduced to simplify calculation of phenomena that are currently computationally or mathematically intractable in their full complexity (McMullin, 1985). They are justified pragmatically: they make tractable what would otherwise be an unsolvable problem. The goal, however, is to eliminate idealization at a future time: "advances in computational power and mathematical techniques should lead the Galilean idealizer to de-idealize, removing distortion and adding back detail to her theories" (Weisberg, 2007, p. 641).

TSMs share the pragmatic justification of Galilean idealizations. They provide an accessible path to understand discrimination tasks that would otherwise be incomprehensible using only an ML model. However, unlike Galilean idealizations, the simplicity of a TSM is not due to a lack of computational power or of mathematical techniques that impedes an accurate account of the ML model. It is quite the opposite. The complexity of the mathematical operations involved in the target system is what makes it impossible to offer a true description of the original model. There is no future time at which a TSM will be de-idealized to match the original ML model. If it were de-idealized, it would become the very object it is meant to explain. Therefore, TSMs are not Galilean idealizations.

Aristotelian idealization is the practice of building models that include only the minimal true causal factors that give rise to a phenomenon. It involves stripping away all properties that are believed to be irrelevant to the problem at hand. Strevens (2008, 2017) offers an analysis of idealization that falls under this category. In his view, an idealization is the omission or distortion of causal factors that are explanatorily irrelevant. These factors might be relevant in the explanation of *how* an event occurred, but they are irrelevant if we want to understand *why* the event occurred. "To understand why a phenomenon obtains, then, is to get a grip on a certain difference-making structure" (Strevens, 2017, p. 42). Idealizations simply ignore apparent difference-makers that were expected to make a difference but did not, and they consequently make it easier to understand the phenomenon. Importantly, Aristotelian idealizations must be factive, i.e., the chosen aspects must be the true causes of the phenomenon.

It is impossible to think of toy surrogate models as Aristotelian idealizations. The main reason is that building a minimal model that preserves some degree of facticity, as Strevens demands, requires finding the minimal set of difference-makers for the prediction or classification task at hand. However, the methods described in the previous section to create TSMs will generate multiple, equally accurate models, as Letham et al. (2015) explain:

> Interpretable models are generally not unique (stable), in the sense that there may be many equally good models, and it is not clear in advance which one will be returned by the algorithm. For most problems, the space of high-quality predictive models is fairly large (…) so we cannot expect uniqueness (Letham et al., 2015, p. 1366).

The existence of multiple equally accurate TSMs has been called the "Rashomon Effect" in ML (Breiman, 2001; Semenova et al., 2022). This multiplicity of surrogate models is possible because there is no known ground truth to which they must respond. Different surrogate models will use different weights and features to successfully complete the prediction or classification task and some of them might replicate spurious correlations present in the original model. In contrast, the causal difference-making structure that a minimal Aristotelian model must capture does not allow for this multiplicity of perspectives. Given an epistemic context, either an aspect of the causal history of a physical or social phenomenon makes a difference or it does not. Uniqueness and factivity are non-negotiable from this perspective.

More generally, the Rashomon Effect also impedes surrogate models from being evaluated in informational and structural terms only. The absence of a ground truth makes it impossible to establish the *correct* representational connection based solely on the features and transition functions that surrogate models supposedly share with their target systems. More importantly, the undeniable *epistemic value* of surrogate models cannot stem only from their informational and/or structural alignment with the target model. This result indicates that understanding the nature of surrogate models requires adopting a pragmatic perspective.[9]

The pragmatic approach to the relation between surrogate and target models adds to the picture the stakeholders and the intended use of the latter. Representation in this approach is seen as an epistemic accomplishment of the users of the model, and it is dependent on their epistemic and practical goals. An examination of how toy models are employed in the engineering sciences and in the formulation of public policies will help clarify how these models can be analyzed from a pragmatic perspective. In these domains, linear models are used to solve problems that would be unnecessarily complex otherwise. Choosing which parameters to include in the model is often a decision guided by practical, not theoretical needs. In particular, parameter selection is based on the conditions of the environment in which the artifact will be deployed. Even the most refined models in engineering and public policy will purposefully leave out known and tractable variables that do not affect the desired level of accuracy and would make the system too cumbersome to implement or to build and manipulate.[10]

Boon and Knuuttila use the phrase "epistemic tools" to describe these simple models:[11]

> From the functional perspective, rather than trying to represent some selected
> aspects of a given target system, modelers often proceed in a roundabout way,

---

[9] This same argument applies to Swoyer's (1991) analysis of representation in structural terms.

[10] Wendy Parker (2020) calls this approach to modeling an adequacy-for-purpose view of model evaluation.

[11] See also Currie (2017) and Knuuttila (2011). Other authors had previously referred to models as "tools" or "instruments" (Cartwright et al., 1995; Morgan & Morrison, 1999). However, as Keller (2000) argues convincingly, what they had in mind was that models can be tools for the construction of better theories; they were not interested in models as pragmatic tools in the sense intended here.

seeking to build hypothetical model systems in the light of their anticipated results or of certain general features of phenomena they are supposed to bring about. If a model gives us certain expected results or replicates some features of the phenomenon, it provides an interesting starting point for further theoretical and experimental conjectures (Boon & Knuuttila, 2009, p. 702).

In the context of ML, TSMs often have a structure that allows for counterfactual or hypothetical reasoning using the main features of the target model. The purpose of TSMs is not to help design or create artifacts, but rather to provide sufficient understanding of the target model to use it and manipulate it. Manipulation is important because algorithmic transparency is increasingly associated with actionability (Longo et al., 2024). For example, a TSM should provide the users of an automated decision system with valuable insights that allow them to make changes that improve their chances of altering the decision of the system. What counts as sufficient understanding of the target model will depend on the epistemic needs of its users and stakeholders, as I explain below. This approach has the advantage of offering a principled reason to adopt a given level of complexity for a surrogate model. In particular, it justifies the use of very simple surrogate models in certain contexts, even if accuracy is partially sacrificed.

But conceiving of toy surrogate models as epistemic tools still does not offer an entirely satisfactory answer to the question about their representational nature. There is obviously no isomorphism or structural similarity[12] between the surrogate model and the target model because they might not even be the same type of algorithm and there is no guarantee that they are using exactly the same features. Pragmatic theories of representation (Suárez, 2004; Giere, 2004, 2010) are based on the idea that the directionality of representation is created by the intentionality involved in the model's intended purpose. In the case of ML, the intentions of the developer of the TSM create the directionality needed to establish a representational relation: a surrogate model is being used and/or interpreted as a model of something else for an intended user, which makes the representative relation triadic, involving human agency.

---

[12] There are well-known objections to the idea that either isomorphism or similarity captures the nature of representation (Suárez, 2003). These objections, which I cannot discuss here, are commonly used to argue for the pragmatic view.

However, intentionality and purpose only establish the directionality of representation; the source of the *epistemic value* of surrogate models must be sought elsewhere. As Tarja Knuuttila points out about models in general,

> If representation is grounded primarily in the specific goals and the representing activity of humans as opposed to the properties of the representative vehicle and its target, nothing very substantial can be said about it in general. This has been explicitly admitted by proponents of the pragmatic approach (cf. Giere 2004), of whom Suárez (2004) has gone farthest in arguing for a "deflationary", or minimalist, account of representation that seeks not to rely on any specific features that might relate the representational vehicle to its target. The minimalist approach has rather radical consequences in terms of how the epistemic value of models should be conceived of. Namely, if we attribute the epistemic value of a model to its being a representation of some target system and accept the minimalist pragmatic notion of representation, not much is established about how we can learn from models (Knuuttila, 2010, pp. 144-145).

The proposal that I want to develop in the rest of the paper is that the epistemic value of TSMs, which can be fleshed out in terms of the pragmatic understanding they provide to situated users in specific contexts, stems from their role in enabling the successful use and manipulation of the target model. Similar ideas can be found in the literature on scientific models. Potochnik (2017), for example, argues that in many cases the commonality sought between representations and what they represent can be understood in terms of functional similarities. The similarities included in the model depend on the functions of interest to the modelers, that is, on the purpose for which they want to use it. Suárez's inferential conception of scientific representation is based on the idea that a model "allows competent and informed agents to draw specific inferences" (Suárez, 2004, p. 773) about a target phenomenon. Presumably, the intended purpose of the model determines the kinds of inferences the agents will be able to draw, and the degree of understanding obtained from the model can be determined by the quality and correctness of their inferences. Finally, in the context of experimental science, Evelyn Fox Keller's (2000) distinction between "models of" and "models for" also emphasizes the idea that models can be analyzed beyond their structure and informational content to include the practical effects that can be derived from their deployment.[13]

To make sense of the multiple ways in which a target model can be used and manipulated, we can use the taxonomy of stakeholders in the ML ecosystem[14] set forth by Tomsett et al. (2018). The authors introduce six kinds of stakeholders according to their role in the ecosystem. To illustrate their taxonomy, I will use the model introduced by Bastani et al. (2017) to classify patients as high or low risk for type II diabetes, examined in the previous section. The first kind of agents in this medical AI ecosystem are the *creators*, the developers and trainers of the random forest. Creators also include system administrators tasked with maintaining and fine-tuning the implemented system. Stakeholders that provide the system's inputs and receive and review its outputs are called *operators* and *executors*. The latter use the outputs to make medical recommendations to the patient. The same person can be both operator and executor, for example, when a medical doctor queries the system with the patient's medical data and makes decisions based on the system's responses. The fourth stakeholder is the patient, the *decision subject*. To train the model, the system was fed with information taken from multiple *data subjects*. And finally, the system's *examiners* are "agents tasked with compliance/safety-testing, auditing, or forensically investigating a system" (Tomsett et al., 2018, p. 6). Examiners operate within a legal framework that includes data privacy regulations, anti-discrimination laws, and medical safety procedures. With the exception of data subjects, all the agents use the system and interact with it in different ways.

Each of these five users has different epistemic goals.[15] *Creators* are interested in debugging the system and improving its performance. They aim to optimize several metrics including predictive accuracy, computational or data efficiency, and bias minimization (Tomsett et al., 2018, p. 5). *Operators* want to make sure the medical data they input to, or the questions they ask of the system are the right ones for it to provide useful information to the executor. *Executors* want to make good evidence-based decisions that consider patients' preferences and are based on sound medical practices. *Decision subjects* want to understand how actionable changes in their behavior, captured by some of the system's features, can lower their risk of diabetes. They also want to be able to trust the recommendations provided by the system to aid the executor's decisions. Finally, *examiners* want to know whether the features used in the model introduce unlawful or unethical criteria in the decision. They will also want to understand the optimization metrics used by creators to make sure the system is safe.[16]

---

[13] Ratti (2020) extends Keller's ideas to the use of models in biological research.

[14] A machine learning ecosystem includes the system and the agents that have interactions with, or are affected by, this system.

[15] The agents' epistemic goals in an ML ecosystem will also depend on other factors such as differences in background knowledge, interests, abilities, and skills. For clarity and brevity, I will ignore those factors here.

Now, given the differences in their epistemic goals, which stakeholders will benefit most from understanding the system globally with the aid of TSMs? Before we tackle this question, it is necessary to explore the connection between understanding and TSMs. We now turn to this question.

## 5. Toy Surrogate Models and Objectual Understanding

An influential definition of interpretability in ML states that it is the "ability to explain or to present in *understandable* terms to a human" (Doshi-Velez & Kim, 2017, p. 2, emphasis added). Another influential author defines interpretability as "the degree to which an observer can *understand* the cause of a decision" (Miller, 2018, p. 8, emphasis added). These definitions, and many others in the ML literature, simply kick the can down the road awaiting a philosophical analysis of understanding in ML. In this section, I present an outline of the type of understanding that can be achieved via TSMs. My goal is to show that these models provide non-factive objectual understanding of the target system and that the understanding they provide is a pragmatic achievement based on the successful use of the target model.

In previous work (Páez, 2019), I have argued that a useful way of looking at the difference between local and global methods in XAI is to classify them according to the kinds of questions they can answer. Local interpretation methods provide answers to *why*-questions: why did this model classify this input as belonging to that class? Global surrogate models, on the other hand, allow agents to answer questions about *what* the system is doing, about its general functioning.

---

[16] Zednik (2021) also uses Tomsett et al.'s taxonomy to explore the explanatory needs of the agents in an ML ecosystem. He examines the types of explanatory questions that the members of the ecosystem are likely to raise and the epistemically relevant elements they need in order to answer them. Zednik focuses only on post hoc interpretability methods and feature detectors, not on surrogate models, which makes this paper a good complement to his approach.

*Prima facie*, understanding a specific decision of the system—through a local interpretation method—and understanding an ML model as a whole—using a decision tree, for example—are two states that demand different accounts of understanding. The former seems to correspond to a form of *understanding-why* while the latter to *objectual understanding* (Páez, 2019). Both types of understanding have been widely discussed in epistemology and the philosophy of science. Let us examine each one in turn.

## 5.1 Objectual Understanding

The characterization of objectual understanding in the epistemological literature fits well with the purpose of TSMs. Zagzebski, for example, argues that understanding is often not directed at discrete propositions in the way knowledge is; rather, it "involves grasping relations of parts to other parts and perhaps the relation of parts to a whole" (Zagzebski, 2009, p. 144). The kinds of relations she has in mind can be spatial, temporal, or causal. "It seems to me that one's mental representation of the relations one grasps can be mediated by maps, graphs, diagrams, and three-dimensional models in addition to, or even in place of, the acceptance of a series of propositions" (p. 145). For Grimm, understanding a complex object such as the New York subway system is a case of knowing-how:

> If know-how implies an apprehension of how a thing works, then it seems to follow that the object of the know-how must be constituted by a structure that can be worked—that is, that can be worked to determine how the various elements of the thing relate to, and depend upon, one another (Grimm, 2011, p. 86).

Both accounts assume that objectual understanding requires being able to identify the various parts of the object, describe their functional interdependence, and use that information to make useful inferences. This is precisely what a TSM offers, either through association rules or directly visible on a decision tree. It aims at providing a simplified version of the original model by providing a view of its main features, of the possible functional interactions between them, and of the transition function between inputs and outputs. Once the user has understood the functional relations of interest for his epistemic needs, as described in the previous section, it can be said that the model has become transparent to him. Transparency is a success concept that depends on being able to grasp the functional structure of the target model through the TSM.

Now, how can we establish that an agent has "grasped" how the model works? A useful way of thinking about the conditions of satisfaction for grasping the interconnectedness of different facts is in terms of an agent's success in *using* the information. This idea has been defended in different guises by philosophers of science like Ylikoski (2009), De Regt (2017), and Kuorikoski (2011, 2023). In the inferential conception of understanding defended by Kuorikoski and Ylikoski (2015), for example,

> [Understanding] is not only about learning and memorizing true propositions, but about the capability to put one's knowledge to use. To understand is to be able to tell what would have happened if things had been different, what would happen if certain things were changed, and what ways there are to bring about a desired change in the outcome (Kuorikoski & Ylikoski, 2015, p. 3819).

More specifically, understanding can be equated "with the ability to draw correct counterfactual what-if inferences about the object of understanding. … To understand a phenomenon is to be able to correctly situate it within a space of possibilities" (Kuorikoski, 2023, p. 218). De Regt also argues that genuine understanding manifests itself as a skill: an agent must have the ability to use his knowledge. Among the most important uses of knowledge in science is the construction of simple, idealized models of some complex phenomenon, which serve as mediators between abstract theories and empirical data. The construction of a successful model with the right idealizing assumptions requires an understanding of the interconnectedness of the data and of the way in which the theory can be applied to the model. Successfully putting one's knowledge to use is not limited to reasoning counterfactually and building models. Being able to fix or improve a system, to profit from it or to game it, are examples of the many possible ways in which usage is a sign of understanding.[17]

Finally, this approach also has the advantage that a person's understanding of a phenomenon or of a subject matter is empirically verifiable. Ylikoski puts the matter thus:

> When we evaluate somebody's understanding, we are not making guesses about his or her internal representations, but about the person's ability to perform according to set standards. The concept of understanding allows that the ability can be grounded in various alternative ways, as long as the performance is correct. Furthermore, the correctness of the internal model is judged by the external displays of understanding, not the other way around. This makes understanding a behavioral concept (Ylikoski, 2009, p. 102).

Having intersubjective criteria for a person's understanding has the advantage of providing ways to test the effectiveness of different models, methods and devices that aid with understanding a phenomenon. Since these criteria are contextual, this approach also allows for the design of epistemic tools that aid different users and populations (Lage et al., 2019). This is essential when we think of the different stakeholders involved in the use of machine learning systems.

In sum, if a necessary condition for having an objectual understanding of a system is to have the ability to use and manipulate it, and to reason counterfactually about it, then the epistemic value of TSMs resides in their potential to make it easier for an agent to perform those tasks with respect to a target model.

More precisely, two conditions must be fulfilled to say that an agent has objectual understanding of a target model. First, the agent must understand how the model maps input states onto corresponding output states, that is, the agent must recognize the main features of the model and identify the transition function $f$. This can be assessed, for example, by probing whether the agent grasps how specific combinations of input variables influence the output of the system. This knowledge will allow the agent to see which variables are more relevant to achieve his or her epistemic goals. Secondly, the agent must be able to match the input and output states, and the transition function, to recognizable features and processes in the real-world learning environment via folk or expert theories.[18] TSMs can help agents in both tasks by highlighting the main variables in the model and graphically or logically illustrating their interdependence via a simpler transition function $f'$. To be helpful to agents in achieving their epistemic goals, TSMs must fulfill two conditions: (i) they must approximate the target model's performance to a reasonable degree, and (ii) their interpretability must be tested via user studies. Without the latter they cannot be catalogued as toy models.

---

[17] An additional pragmatic aspect of this approach is that there is no unique benchmark for understanding. In De Regt's (2023) view, understanding is contextual. Criteria for understanding and intelligibility depend on the historical and disciplinary context.

Let us return now to the diabetes risk assessment model and its ecosystem of users to illustrate the previous analysis. Which of them can benefit from the global understanding provided by the decision tree extracted from the opaque model? Given their epistemic goals, described at the end of the previous section, it seems evident that creators, decision subjects, operators, executers, and examiners can all benefit from using the TSM. For brevity, I will limit the analysis to the first two. Creators can examine the toy model to improve performance metrics. "As long as the approximation quality is good, then the interpretable model mirrors the computation performed by the complex model. Thus, by inspecting the interpretable model, the data scientist can diagnose issues in the complex model" (Bastani et al., 2019, p. 1).[19] Creators are also uniquely situated to provide surrogate models to the other members of the ecosystem to help them achieve their epistemic goals. Decision subjects, on the other hand, obtain the highest benefits from the decision tree. They can use the information on the decision tree to make significant changes to their lifestyles. For example, the tree shows that the combination of lower cholesterol levels and lower tobacco consumption significantly affects whether a patient will be classified as high or low risk. Other variables such as age and hypothyroidism are not actionable, but they increase risk levels when combined with other actionable variables. Decision trees also have the advantage of showing "rosier" scenarios in which a change of habits and lifestyle can lead to a better outcome.

---

[18] Zednik (2021) argues that these two elements correspond to Marr's (1982) computational level of analysis, which centers on *what* a system does and *why* it does what it does. Specifying why a system does something in Marr's sense is not the same as answering why-questions in the epistemological sense discussed in the next section. Rather, it is a global specification of the representational content of the input and output variables and of the transition function within the learning environment.

[19] Naturally, surrogate models are just one of several XAI tools available to creators. In particular, local post hoc interpretation methods continue to be important for creators, who want to make sure that the model is using the "right" features. Furthermore, creators will also want to intervene at the algorithmic and implementation levels, to use Marr's analytical terminology, a purpose for which surrogate models are clearly inadequate.

Prima facie, this description of how decision subjects can use TSMs to understand the risk assessment score generated by a black box model sounds like a case of counterfactual reasoning, along the lines of the method introduced in Wachter et al. (2018). In the last section of the paper, I will show that a successful use of such counterfactual methods always depends on having a previous objectual understanding of the system.

## 5.2 Understanding-Why

The epistemological accounts of understanding-why fit well with local *post-hoc* interpretability methods. Understanding why *p* is not equivalent to simply *knowing* why *p*. Knowing that an image recognition system correctly classified an image as a dog because it was presented with an image of a dog is clearly not enough to understand the decision. The person must be able to answer a wide range of questions of the type what-if-things-had-been-different (Woodward, 2003, p. 221). What if the ears of the dog were not visible? What if the light had been weaker? What if the input were the mirror image of the original? Local *post hoc* interpretation methods allow users to visualize variations of the input in ways that provide answers to these questions, especially when users are allowed to query the system.

Many authors have argued that not just objectual understanding but understanding *in general* requires the ability to envision different configurations of the parts of an object and infer its resultant states. In other words, they claim that understanding requires the ability to think counterfactually (de Regt & Dieks 2005; Wilkenfeld 2013). "Understanding of the possible is the way to understand why the actual emerged and how it functions" (Knuuttila, 2011, p. 269). But appearances to the contrary, local *post-hoc* interpretability methods do not provide genuine counterfactual understanding. By tinkering with the input, users can only establish piecemeal correlations that cannot be generalized in any obvious way.[20] What prevents users from engaging in genuine

---

[20] To be sure, developers can use these correlations as cumulative *evidence* to infer, via induction, a hypothesis about the main features responsible for the system's decisions. However, a global

counterfactual reasoning is the lack of functional information about the system, that is, the lack of objectual understanding. TSMs provide general rules that have been mined from the data or extracted directly from the model, thus providing the underlying functional scaffolding required to reason counterfactually. One can follow one branch or another of a decision tree, and each path will be a counterfactual case that will be entirely determined by the static functional structure depicted in the tree.

Naturally, knowledge of the domain of application of the model reinforces our objectual understanding. In particular, the empirically validated theoretical knowledge that determined the criteria utilized to select the training and testing samples, and the labels used to classify the data will strengthen our understanding of the model and our trust in its predictions.[21] In a similar vein, Sullivan (2022) argues that an opaque ML system can provide understanding if there is adequate scientific evidence supporting the link connecting the model to the target-phenomenon. Such evidence is used to validate the findings of local *post-hoc* methods such as saliency maps. Thus, she also seems to believe that local *post-hoc* methods by themselves are insufficient to provide understanding of the decisions of the system. Understanding-why always requires some degree of objectual understanding (Páez, 2019).

Karimi et al. (2020) offer an argument in a similar direction but from a more practical perspective. They focus on the problem of algorithmic recourse. When a person has been affected by an unfavorable automated decision (e.g., a rejected loan application), so-called nearest counterfactual explanations (Wachter et al., 2018) can be used to suggest actions that a person can take to achieve a favorable decision from an ML system. The authors show that such counterfactuals "do not translate to an *optimal* or *feasible* set of actions that would favorably change the prediction of *h* if acted upon. This shortcoming is primarily due to the lack of consideration of causal relations governing the world" (Karimi et al., 2020, p. 359). The missing causal information is part of the theoretical knowledge included in the agent's objectual understanding of the system. To be sure, TSMs by themselves do not provide the missing causal information, but they at least

---

explanation provides a more systematic way of establishing these features and the relationships between them. I thank an anonymous reviewer for pointing this out.

[21] Ribeiro et al. (2016), creators of LIME, seem to acknowledge this when they say that trust in a prediction presupposes "prior knowledge about the application domain" (p. 1136).

provide human-interpretable functional correlations that can be empirically investigated and validated. They can be seen as an initial step towards obtaining the causal knowledge required to design actionable decision systems.

In sum, local *post hoc* interpretability methods *by themselves* cannot provide understanding of a model.[22] By providing the causes of specific predictions of the model these methods can contribute to establish the interconnections between features and outputs, but only an agent's ability to reason counterfactually about the model, only her ability to use it and manipulate it, can be taken as evidence that she has understood it, that it has become transparent to her. TSMs are perhaps the best epistemic tool available for a wide variety of stakeholders to master the workings of opaque target models.

## 5. Conclusion

In this paper I have explored the nature and epistemic role of TSMs in ML. An essential difference with most toy models in the sciences is that the representational relation between the model and its target can only be understood in terms of its role in enabling the successful use and manipulation of the target model, which are signs of pragmatic understanding. They cannot be seen as Galilean idealizations because of their non-transient nature, nor as Aristotelian idealizations because of their non-factive nature. In general terms, any relation based solely on the informational and structural content of the models will fail to account for the epistemic value of the TSM. By focusing on their role as epistemic tools that provide pragmatic understanding, it is easier to account for their epistemic value. I have also argued that, from an epistemological point of view, objectual understanding of the target model should precede the use of local *post-hoc* interpretability methods because the global understanding of the model is required to answer why- and how-possibly questions about its individual predictions. Lastly, I hope to have established that TSMs are a new object of study for theories of understanding, one that is not easily accommodated under current analyses.

## Acknowledgments

For helpful comments on previous versions of this paper, I would like to thank David Danks, Atoosa Kasirzadeh, Emanuele Ratti, and audiences at the 2022 SURe Conference at Fordham University in New York, at the workshop: Issues in XAI #5: Understanding Black Boxes, Interdisciplinary Perspectives, in Dortmund, and at the ALFAn VI conference in Santiago de Chile.

---

[22] I develop this argument in much more detail in Páez (2019).